# Personalized Scorer Modeling: A Learning-Based Framework for Deriving Robust Sleep Stage Labels from Multiple Experts

Seyyed Ali Hoseini[1]*, Javad Baseri[1], Hamid Saadatfar[1], Edris Hoseini Gol[1], AmirHossein Eshghi[1]

1 Department of Electrical and Computer Engineering, University of Birjand, Birjand, South Khorasan, Iran

* Corresponding author: sa.hosseini@birjand.ac.ir

## Abstract

Sleep stage classification is central to the diagnosis, monitoring, and treatment of sleep-related disorders. Most automatic sleep staging studies evaluate models against a single reference hypnogram, although clinical sleep staging is affected by inter-scorer variability and no individual scorer can be considered error-free. Multi-scored datasets, in which the same polysomnographic recordings are annotated by several experts, provide an opportunity not only to train automatic classifiers but also to derive more reliable target labels from the collective behavior of multiple scorers. This study investigates two publicly available multi-scored sleep datasets, DOD-H and DOD-O. Electroencephalogram signals from the C3-M2 channel and chin electromyogram signals were segmented into 30-second epochs. For each signal, 25 frequency-domain features and 5 time-domain features were extracted, producing 30 features per modality and 60 features when EEG and EMG were combined. The proposed learning-based hypnogram (LBH) is generated by modeling the stage-specific behavior of each scorer through confusion matrices obtained from machine-learning models. These confusion matrices are column-normalized to estimate the probability of each true sleep stage given the label assigned by each scorer; the probabilities from all scorers are then aggregated to select the final stage label for each epoch. The generated labels were evaluated using random forest, support vector machine, and multilayer perceptron classifiers under EEG-only and EEG+EMG settings and were compared with the original dataset hypnogram (DH) and the best-scorer hypnogram (BSH). Across the experiments, LBH consistently improved the reported total metrics. The best performance was obtained by random forest with EEG+EMG features and the proposed LBH, reaching 86.07% accuracy, 85.46% precision, and 85.29% F1-score on DOD-H, and 86.04% accuracy, 85.21% precision, and 84.70% F1-score on DOD-O. These results suggest that personalized scorer modeling can improve the construction of reference hypnograms in multi-scored sleep datasets without discarding the information provided by individual experts.

## 1. Introduction

Sleep is a fundamental biological process that affects both mental and physical health. Healthy adults typically require approximately 7 to 9 hours of sleep per day, meaning that humans spend nearly one-third of their lives sleeping. Sleep contributes to energy conservation, memory consolidation, information processing, physical growth, muscle repair, immune function, and overall daily performance [1-3]. Insufficient sleep and sleep-related disorders can impair cognitive ability and productivity and can increase the risk of stress, depression, hypertension, metabolic dysfunction, and other mental and physical disorders [4-11]. Sleep quality is therefore as important as sleep duration, and it is influenced by environmental, psychological, and physiological factors [12]. Reliable assessment of sleep quality is consequently important for clinical diagnosis, treatment planning, and sleep research.

According to the American Academy of Sleep Medicine (AASM), sleep is generally scored into Wake (W), non-rapid eye movement stages N1, N2, and N3, and rapid eye movement sleep (REM or R) [13]. A whole-night sleep recording is represented as a sequence of these stages. The resulting sequence, called a hypnogram, visualizes sleep architecture and the repeated sleep cycles that typically occur during the night [14]. Manual sleep staging is still the clinical reference procedure, but it is time-consuming and depends on expert interpretation of polysomnographic signals. As a result, sleep-stage labels can vary across scorers, particularly for transitional and low-prevalence stages such as N1. Automatic sleep staging uses machine-learning or deep-learning algorithms to classify sleep stages from physiological signals and can support clinical practice, large-scale research, physician training, and trials.

A limitation of many automatic sleep staging studies is that they optimize the classifier against one fixed reference hypnogram. Even if an automatic method reproduces that reference perfectly, it may only reproduce the opinion of a single scorer or of a label-generation rule. In practice, scorers disagree, and no single scorer is free from error. Multi-scored datasets directly expose this variability by providing multiple expert labels for the same epochs. The present study focuses on two multi-scored sleep datasets, DOD-H and DOD-O [15]. In these datasets, polysomnographic recordings were labeled by five experienced scorers. The majority agreement is high for most epochs, but the strength of agreement differs by stage and by scorer. Instead of treating all scorer labels as equally reliable, this paper proposes a personalized scorer modeling framework that estimates stage-specific reliability profiles for each scorer and uses these profiles to generate a learning-based hypnogram.

The main contributions of this study are as follows. First, the agreement structure of the DOD-H and DOD-O multi-scored datasets is organized and reported by sleep stage. Second, a scorer-specific modeling strategy is proposed in which each scorer is characterized through a confusion matrix estimated with machine learning. Third, the scorer-specific matrices are converted into stage-wise probabilities and aggregated to derive a learning-based hypnogram. Fourth, the proposed labels are evaluated across three classifiers, two signal configurations, and two datasets to show that the labeling strategy improves the presentation and reproducibility of the reported results.

## 2. Related Work

A large body of work has investigated automatic sleep stage classification using polysomnographic and wearable signals. Lee et al. [16] used the SHHS, WSC, MESA, NCHSDB, SNUH, CCC, and MNC datasets and developed a deep neural network for sleep staging based on heart rate and limb-movement features, reporting accuracies of 76.6% on CCC and 81.0% on SNUH. Zhou et al. [17] proposed a lightweight segmented attention network and reported accuracies of 85.8%, 86.4%, and 82.5% on Sleep-EDFX, MASS, and HSFU, respectively.

Deep-learning architectures that combine convolutional and recurrent modules have also been widely used. Yan et al. [18] proposed a CNN-LSTM architecture for multimodal time series using SHHS, ISRUC, and Sleep-EDF recordings and reported approximately 87% accuracy on SHHS-1 and 86% on Sleep-EDF and ISRUC. Khalili and Asl [19] proposed a temporal convolutional neural network with a mapping-label data augmentation technique for raw single-channel EEG, reaching 85.39% accuracy on Sleep-EDF-2013 and 82.46% on Sleep-EDF-2018. Michielli et al. [20] proposed a cascaded LSTM recurrent neural network for single-channel EEG and reported 86.7% accuracy on Sleep-EDF.

Several studies have used convolutional or hybrid models with EEG and related PSG signals. Yildirim et al. [21] used EEG and EOG signals from Sleep-EDF and Sleep-EDFX and proposed a one-dimensional convolutional neural network, reporting 91.22% accuracy for five-class classification on Sleep-EDF and 90.98% on Sleep-EDFX. Supratak et al. [22] introduced DeepSleepNet, combining CNNs and bidirectional LSTM layers for single-channel raw EEG, and reported 82.0% accuracy on Sleep-EDF and 86.2% on MASS.

Classical machine-learning approaches remain relevant because they are interpretable and computationally efficient. Sharma et al. [23] used EEG, EOG, and EMG signals from the SHHS-1 and SHHS-2 datasets and introduced a wavelet-based Tsallis entropy sleep scoring system, reporting accuracies of 84.3% and 86.3%, respectively. Satapathy et al. [24] proposed an ensemble stacking model using single-channel EEG and age features, achieving 92.79% accuracy on Sleep-EDF and 83.2% on Sleep-EDF Expanded. Vallat and Walker [25] used multiple cohorts, including DOD-H and DOD-O, and extracted spectral features from EEG, EOG, and chin EMG; with a gradient-boosting classifier, they reported 86.6% accuracy on DOD-H and 84.3% on DOD-O. Ghimatgar et al. [26] extracted time-domain features from EEG epochs and evaluated SVM, LDA, random forest, nearest neighbor, and decision tree classifiers across S-EDF, SE-EDF, ISRUC3, and DRM-SUB.

Abdulla et al. [27] proposed an intelligent model based on multi-channel spectrum-pattern and texture features for automatic sleep stage classification. Their method transformed 30-second EEG segments into spectral images, extracted Multiple channels Information Local Binary Pattern features, and used an ensemble classifier with genetic optimization; their comparison table included accuracies of 93.21% and 92.93% under two scoring settings.

Although these studies have improved automatic sleep classification, relatively few have focused on the problem of how the reference hypnogram should be constructed when multiple scorers label the same recordings. Multi-scored datasets are richer than single-label datasets because they preserve uncertainty, disagreement, and scorer-specific tendencies. The present study addresses this gap by focusing on label generation from multiple scorers rather than on classifier architecture alone.

# 3. Materials and Methods

## 3.1. Datasets and Scoring Protocol

The experiments use two publicly available multi-scored sleep datasets introduced as the Dreem Open Datasets [15]. DOD-H contains polysomnographic recordings from 25 healthy volunteers, while DOD-O contains recordings from 55 patients with obstructive sleep apnea. In both datasets, each recording was divided into 30-second epochs and scored by five sleep technicians with at least five years of experience from three different sleep research centers.

Each scorer assigned one of five sleep-stage labels to each epoch: 0 for Wake, 1 for N1, 2 for N2, 3 for N3, and 4 for REM. Some epochs, particularly at the beginning or end of recordings where movement or wakefulness can produce large signal variations, received an unspecified label (-1). These unspecified or removed epochs were excluded from the final dataset hypnograms, whereas the full scorer annotations remain available in the original data.

**Table 1.** Summary of the multi-scored datasets used in the study.

| Dataset | Population | Subjects | Raw epochs reported | Retained hypnogram epochs used in tables | Scorers |
|---|---|---|---|---|---|
| DOD-H | Healthy volunteers | 25 | 25,441 | 24,662 | 5 |
| DOD-O | Obstructive sleep apnea patients | 55 | 60,371 | 53,233 | 5 |

## 3.2. Agreement Analysis and Existing Dataset Hypnogram

To analyze the behavior of the scorers, all five labels for every epoch were organized together with the dataset hypnogram. The agreement of each scorer with the majority vote was quantified using a soft-agreement measure. For scorer j, the score is the average indicator of whether that scorer assigned the same label as the majority label for epoch i:

$$SA_j = \frac{1}{k}\sum_{i=1}^{K} I(y_{i,j} = MV_i) \tag{1}$$

Here, K is the number of epochs, $y_{i,j}$ is the label assigned by scorer $j$ to epoch $i$, $MV_i$ is the majority-vote label for epoch $i$, and $I(.)$ is the indicator function. A higher soft-agreement score means that the scorer agrees more often with the majority decision across the dataset.

For DOD-H, the soft-agreement scores of the five scorers were 0.87, 0.91, 0.92, 0.84, and 0.92. Therefore, scorers 1, 2, 3, and 5 were selected for the dataset hypnogram rule, while scorer 4 was excluded. For DOD-O, the scores were 0.88, 0.88, 0.87, 0.88, and 0.91. Therefore, scorers 1, 2, 4, and 5 were selected, while scorer 3 was excluded. When the four selected scorers produced a two-versus-two tie, the label associated with the higher-scoring scorer was selected.

**Table 2.** Scorer agreement status in the DOD-H dataset. Percentages are computed within each sleep stage and for the total retained epochs.

| **Measure** | **REM** | **N3** | **N2** | **N1** | **Wake** | **Total** |
|---|---|---|---|---|---|---|
| Number of epochs | 4727 | 3514 | 11879 | 1505 | 3037 | 24662 |
| Percentage | 19.17 | 14.25 | 48.17 | 6.10 | 12.31 | - |
| Agreement of all 5 scorers | 3085 | 1992 | 7227 | 182 | 2023 | 14509 |
| Percentage | 65.26 | 56.69 | 60.84 | 12.09 | 66.61 | 58.83 |
| Agreement of 4 scorers | 1027 | 877 | 2909 | 418 | 558 | 5789 |
| Percentage | 21.73 | 24.96 | 24.49 | 27.77 | 18.37 | 23.47 |
| Agreement of 3 scorers | 502 | 539 | 1613 | 681 | 392 | 3727 |
| Percentage | 10.62 | 15.34 | 13.58 | 45.26 | 12.91 | 15.11 |
| Conditional agreement | 113 | 106 | 130 | 224 | 64 | 637 |
| Percentage | 2.39 | 3.01 | 1.09 | 14.88 | 2.11 | 2.59 |

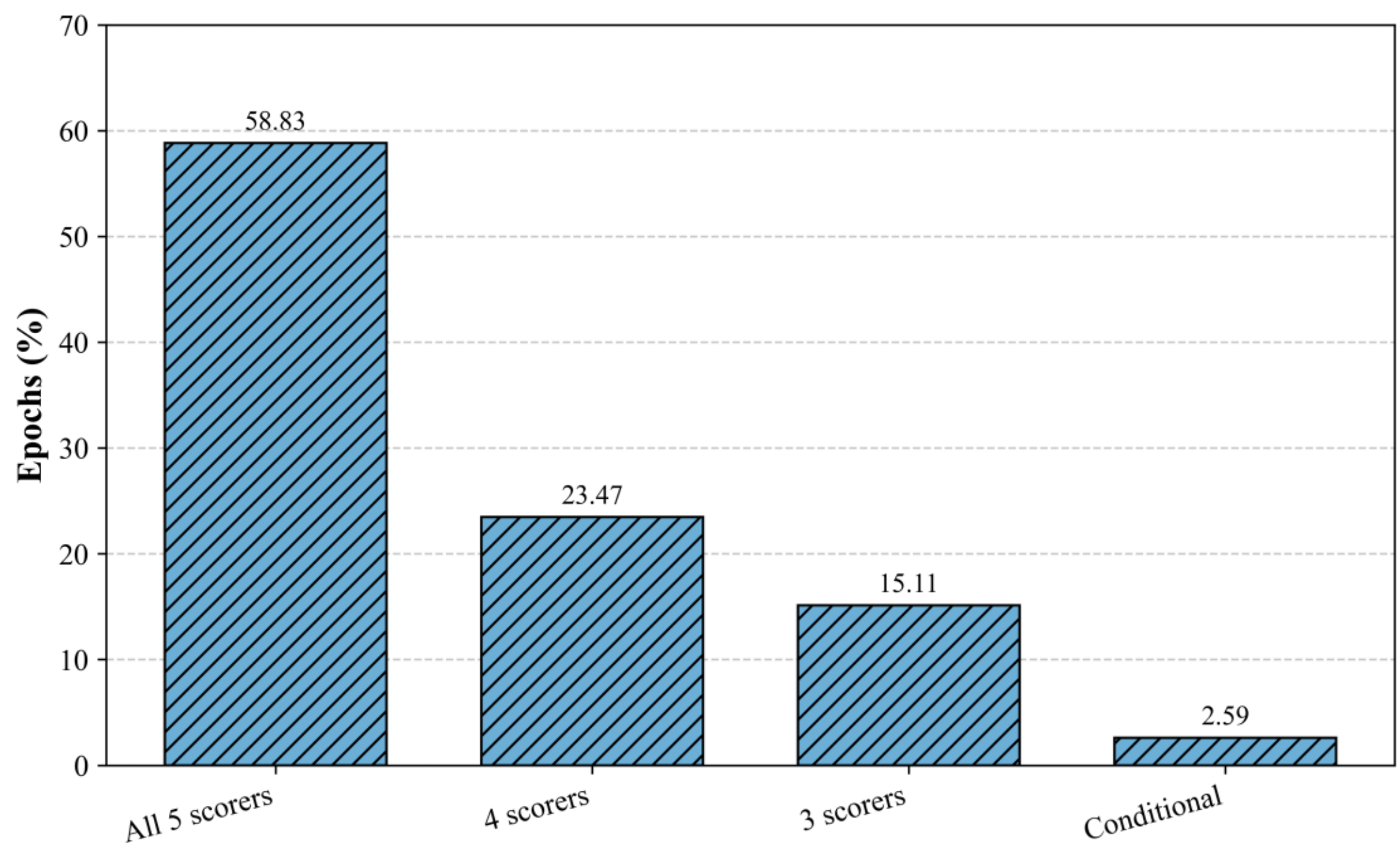

**Figure 1.** Overall agreement distribution of scorers in the DOD-H dataset.

**Table 3.** Scorer agreement status in the DOD-O dataset. Percentages are computed within each sleep stage and for the total retained epochs.

| **Measure** | **REM** | **N3** | **N2** | **N1** | **Wake** | **Total** |
|---|---|---|---|---|---|---|
| Number of epochs | 8103 | 5500 | 26267 | 2860 | 10503 | 53233 |
| Percentage | 15.23 | 10.32 | 49.35 | 5.37 | 19.73 | - |
| Agreement of all 5 scorers | 5675 | 1931 | 14650 | 347 | 7566 | 30169 |
| Percentage | 70.04 | 35.11 | 55.77 | 12.14 | 72.03 | 56.68 |
| Agreement of 4 scorers | 1531 | 1743 | 6118 | 597 | 1655 | 11644 |

| Measure | REM | N3 | N2 | N1 | Wake | Total |
|---|---|---|---|---|---|---|
| Percentage | 18.89 | 31.69 | 23.29 | 20.87 | 15.76 | 21.87 |
| Agreement of 3 scorers | 780 | 1923 | 4744 | 1352 | 1023 | 9822 |
| Percentage | 9.63 | 34.96 | 18.06 | 47.27 | 9.74 | 18.45 |
| Conditional agreement | 117 | -97 | 755 | 564 | 259 | 1598 |
| Percentage | 1.44 | -1.76 | 2.88 | 19.72 | 2.47 | 3 |

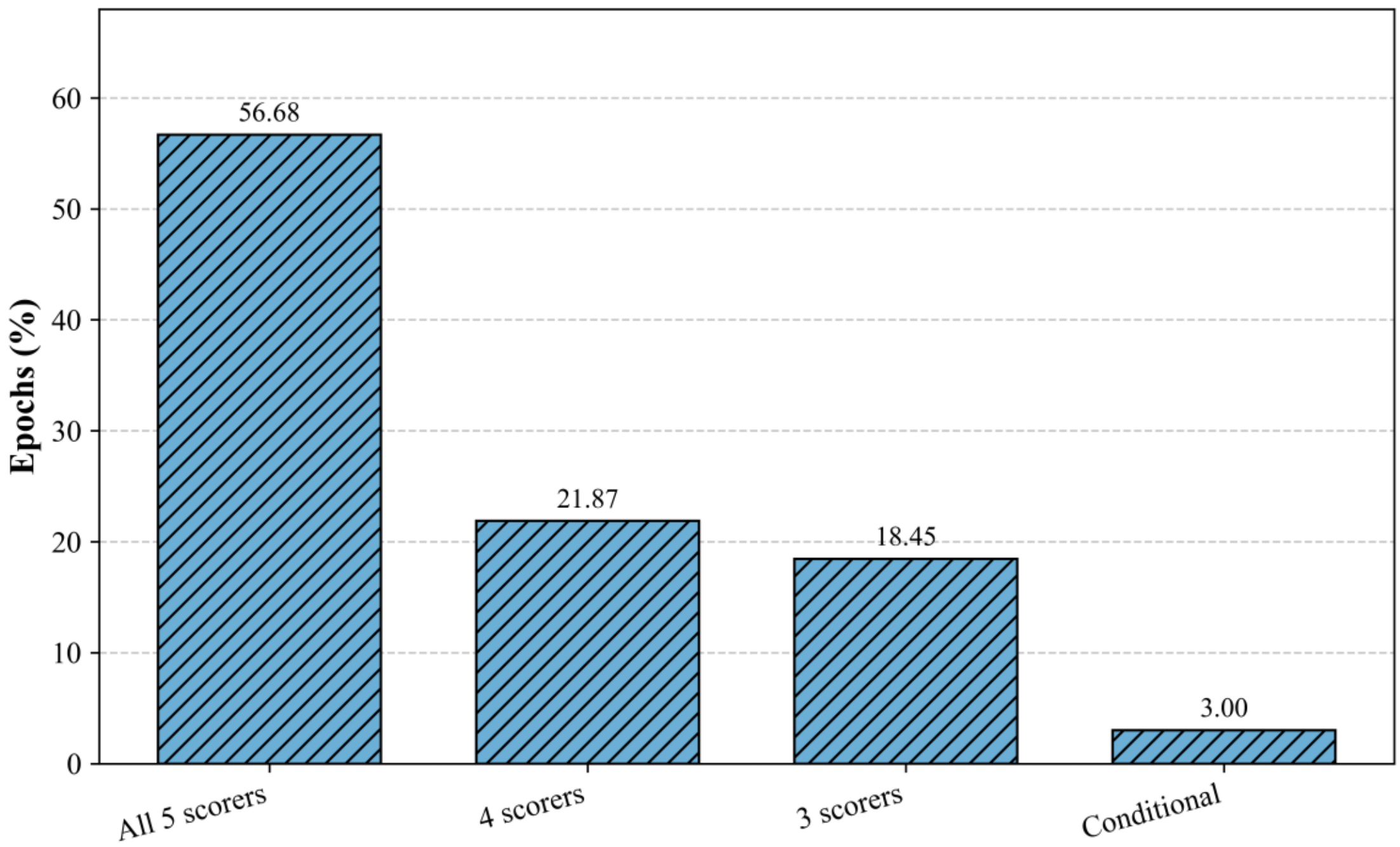

**Figure 2.** Overall agreement distribution of scorers in the DOD-O dataset.

The agreement analysis shows that N1 is the most ambiguous stage in both datasets. In DOD-H, only 12.09% of N1 epochs received agreement from all five scorers, whereas 45.26% had agreement from only three scorers. In DOD-O, the same pattern is observed: 12.14% of N1 epochs received agreement from all five scorers, while 47.27% had agreement from three scorers. This class-specific disagreement motivates the proposed scorer-aware label aggregation strategy.

The negative conditional value reported for N3 in DOD-O arises from the way the distributed dataset hypnogram was constructed after removing one scorer and resolving four-scorer ties. Some originally three-person agreements were converted into two-person agreements or adjusted during the tie-resolution process, reducing the number of N3 labels in that conditional category. This observation is retained because it clarifies a non-intuitive value in the original agreement table.

## 3.3. Signal Channels and Feature Extraction

For each 30-second epoch, features were extracted from the EEG C3-M2 channel and from the chin EMG signal. Each modality contributed 30 features: 25 frequency-domain features and 5 time-domain features. Therefore, EEG-only experiments used 30 features per epoch, while EEG+EMG experiments used 60 features per epoch. Frequency-domain features consisted of frequency power values between 0 and 40 Hz. Ten features were extracted in the 0-10 Hz range with 1-Hz resolution. Ten additional features were extracted in the 10-40 Hz range with 3-Hz resolution. Five conventional band-power features were also extracted for delta (0.5-4 Hz), theta (4-8 Hz), alpha (8-12 Hz), sigma (12-16 Hz), and beta (16-30 Hz). Time-domain features included variance, standard deviation, zero-crossing count, coefficient of variation, and the ratio between the first- and second-order coefficients of variation.

**Table 4.** Feature groups extracted from EEG and EMG signals.

| Feature group | Number per signal | Description | Frequency or time range |
|---|---|---|---|
| Low-frequency power bins | 10 | One power feature for each 1-Hz bin | 0-10 Hz |
| High-frequency power bins | 10 | One power feature for each 3-Hz bin | 10-40 Hz |
| Canonical band powers | 5 | Delta, theta, alpha, sigma, beta power | 0.5-30 Hz |
| Temporal features | 5 | Variance, standard deviation, zero crossings, coefficient of variation, and ratio of first- and second-order coefficients of variation | Time domain |
| Total | 30 per signal | 30 EEG features; 30 EMG features; 60 combined EEG+EMG features | Per 30-s epoch |

## 3.4. Baseline and Proposed Hypnograms

Three hypnogram definitions were evaluated. The best-scorer hypnogram (BSH) uses the labels of the scorer with the highest overall soft-agreement. The dataset hypnogram (DH) is the hypnogram distributed with the DOD-H and DOD-O datasets after scorer selection and tie resolution. The learning-based hypnogram (LBH) is the proposed hypnogram generated by personalized scorer modeling. The key idea of LBH is to use the observed relationship between physiological features and each scorer's labels to estimate a stage-specific reliability profile for every scorer. A scorer can be strong for one stage and weak for another. Therefore, rather than assigning a single global weight to each scorer, the proposed method learns stage-dependent weights from confusion matrices.

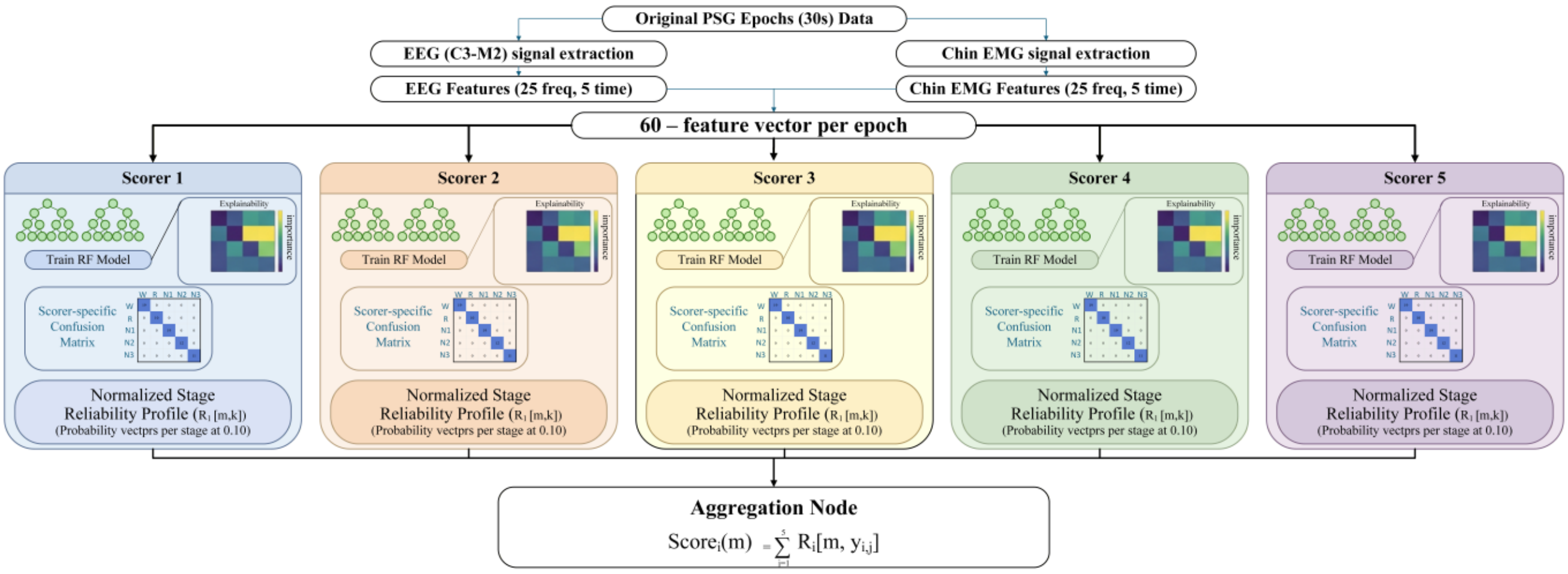

**Figure 3.** Workflow of the proposed learning-based hypnogram generation method.

## 3.5. Personalized Scorer Modeling and LBH Generation

For each scorer $j$, a random forest classifier was trained and tested using five-fold cross-validation. The input consisted of the extracted EEG and EMG features, and the target labels were the labels provided by that scorer. The resulting confusion matrix $C_j$ describes how the model reproduces the scorer's labeling pattern for different sleep stages. Rows of the confusion matrix represent reference sleep stages and columns represent predicted or assigned stages. For each scorer, each column was normalized so that it could be interpreted as the probability of a candidate true stage $M$ given that scorer $j$ assigned or predicted stage $k$. This normalization gives a stage-specific scorer profile rather than a single global scorer weight.

$$R_j[M,K]=\frac{C_j[M,K]}{\sum_{r=1}^{M} C_j[r,K]} \quad (2)$$

In Equation (2), $C_j$ is the confusion matrix for scorer $j$, $M$ is the number of sleep stages, $K$ is the stage label assigned by the scorer, m is a candidate final stage, and $R_j[M,K]$ is the normalized reliability value for scorer $j$. For epoch $i$, the label assigned by scorer $j$ is denoted $y_{i,j}$. The proposed method collects the corresponding reliability column for each scorer and sums the stage probabilities across all $N$ scorers.

$$Score_i(M)=\sum_{j=1}^{N} R_j[M,y_{i,j}] \tag{3}$$

The final LBH label is the stage with the maximum aggregated score.

$$\hat{y}_i^{LBH} = \arg\max Score_i(M) \tag{4}$$

This procedure was applied separately to DOD-H and DOD-O. The generated LBH labels and the hypnograms provided by the five scorers are publicly available at *https://github.com/Edris-Hosseini/dreem-learning-hypnogram.*

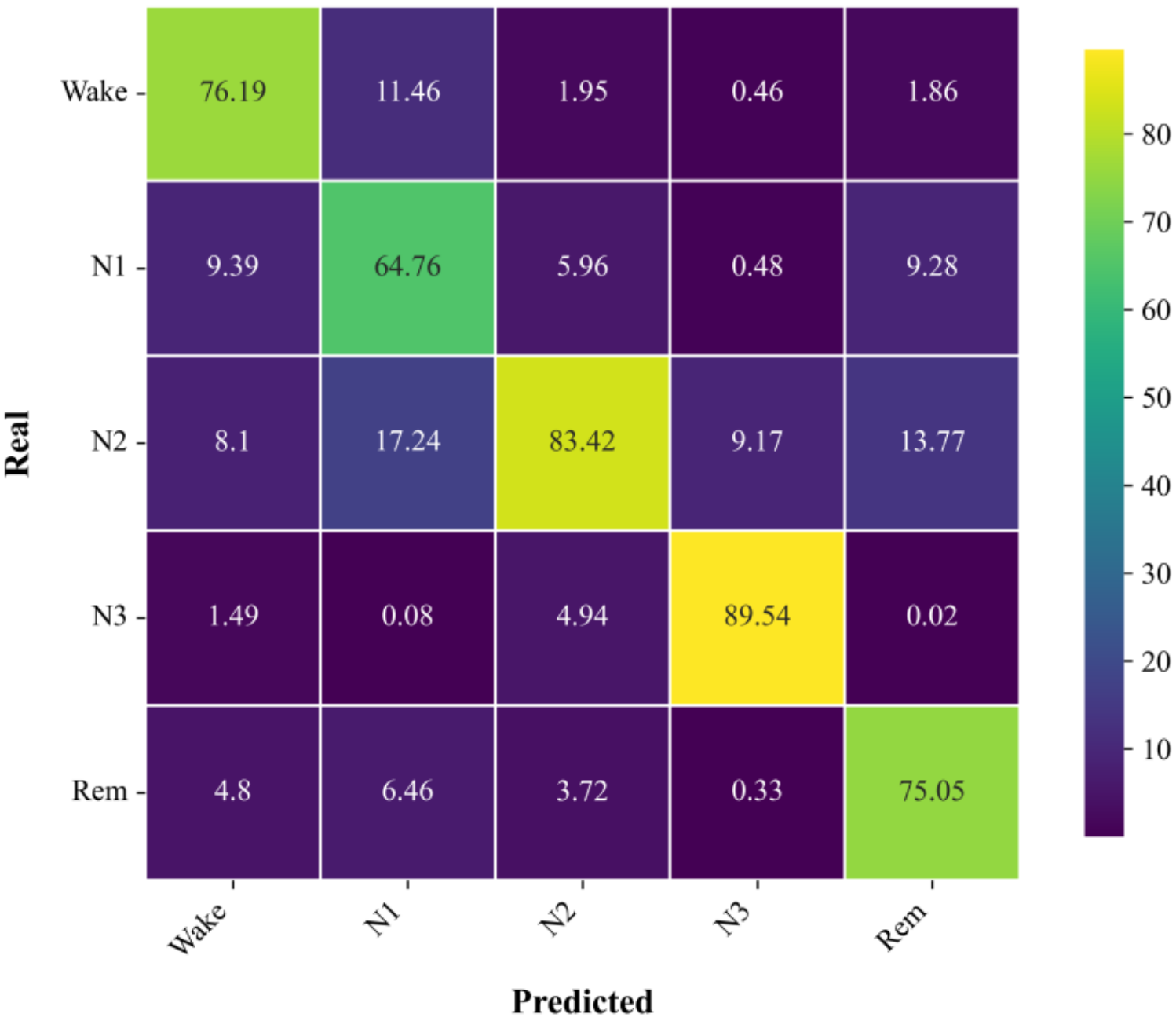

**Figure 4.** Confusion-matrix values for scorer 1 in the DOD-H dataset. Rows represent real stages and columns represent predicted stages.

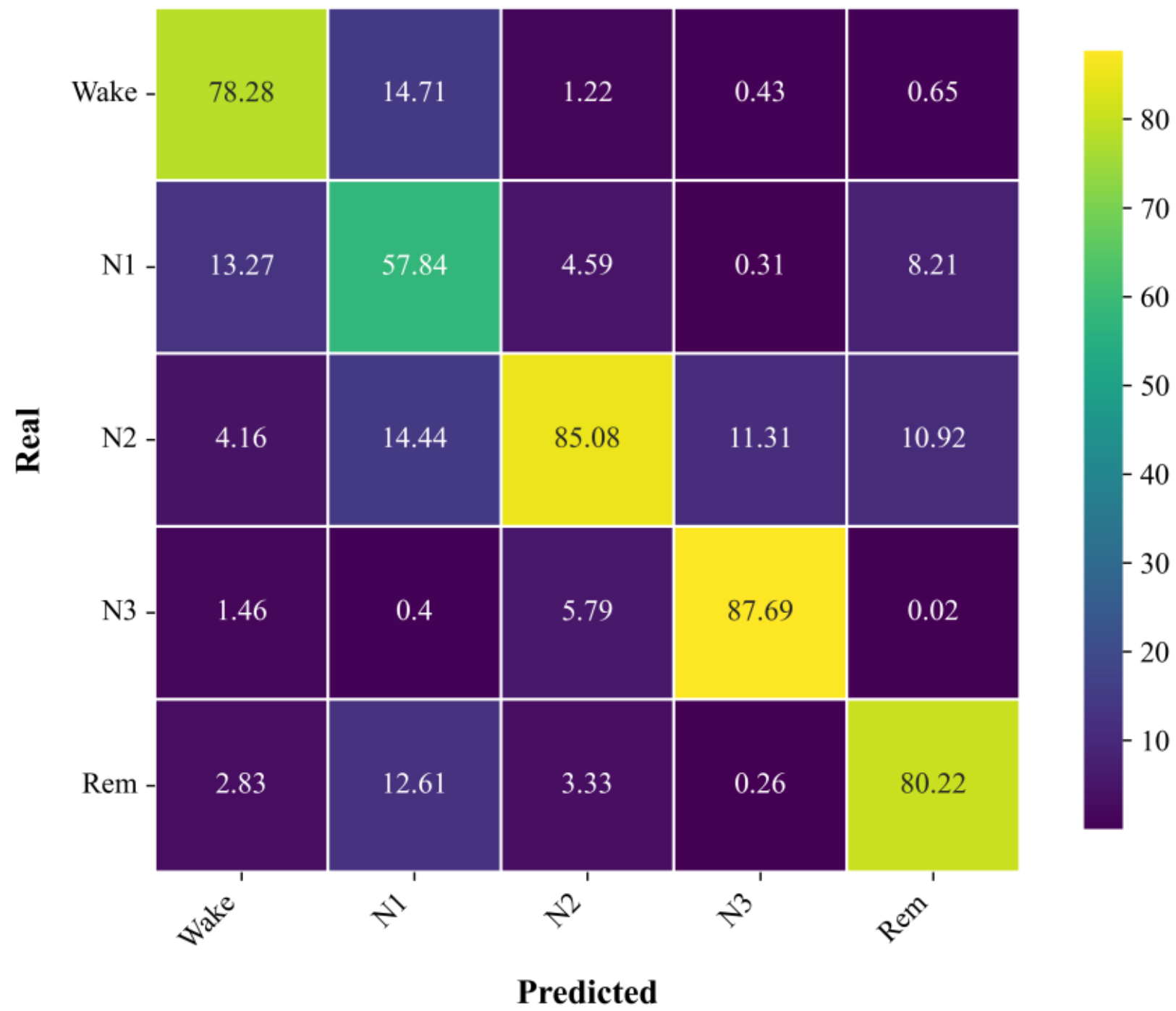

**Figure 5.** Confusion-matrix values for scorer 2 in the DOD-H dataset. Rows represent real stages and columns represent predicted stages.

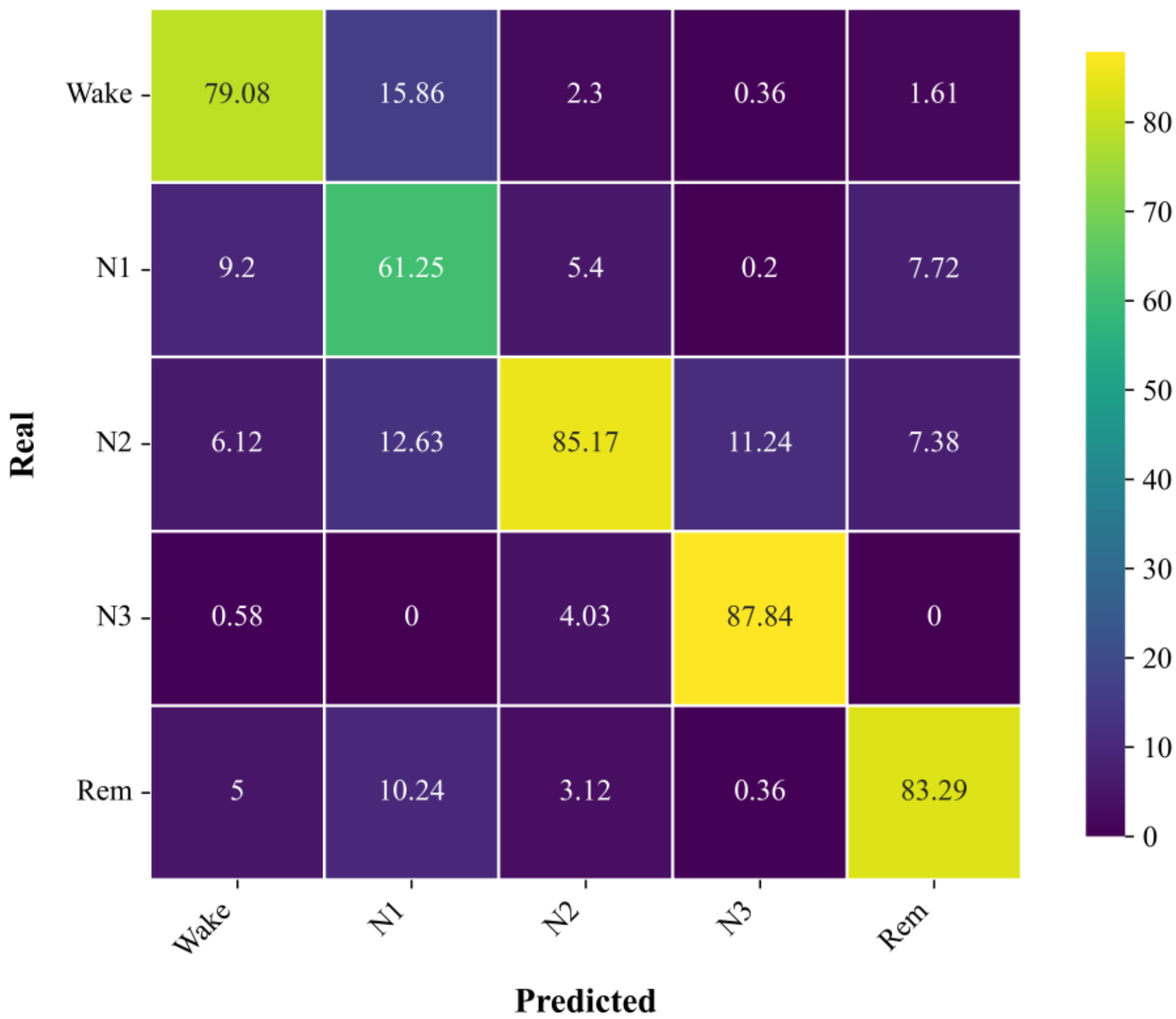

**Figure 6.** Confusion-matrix values for scorer 3 in the DOD-H dataset. Rows represent real stages and columns represent predicted stages.

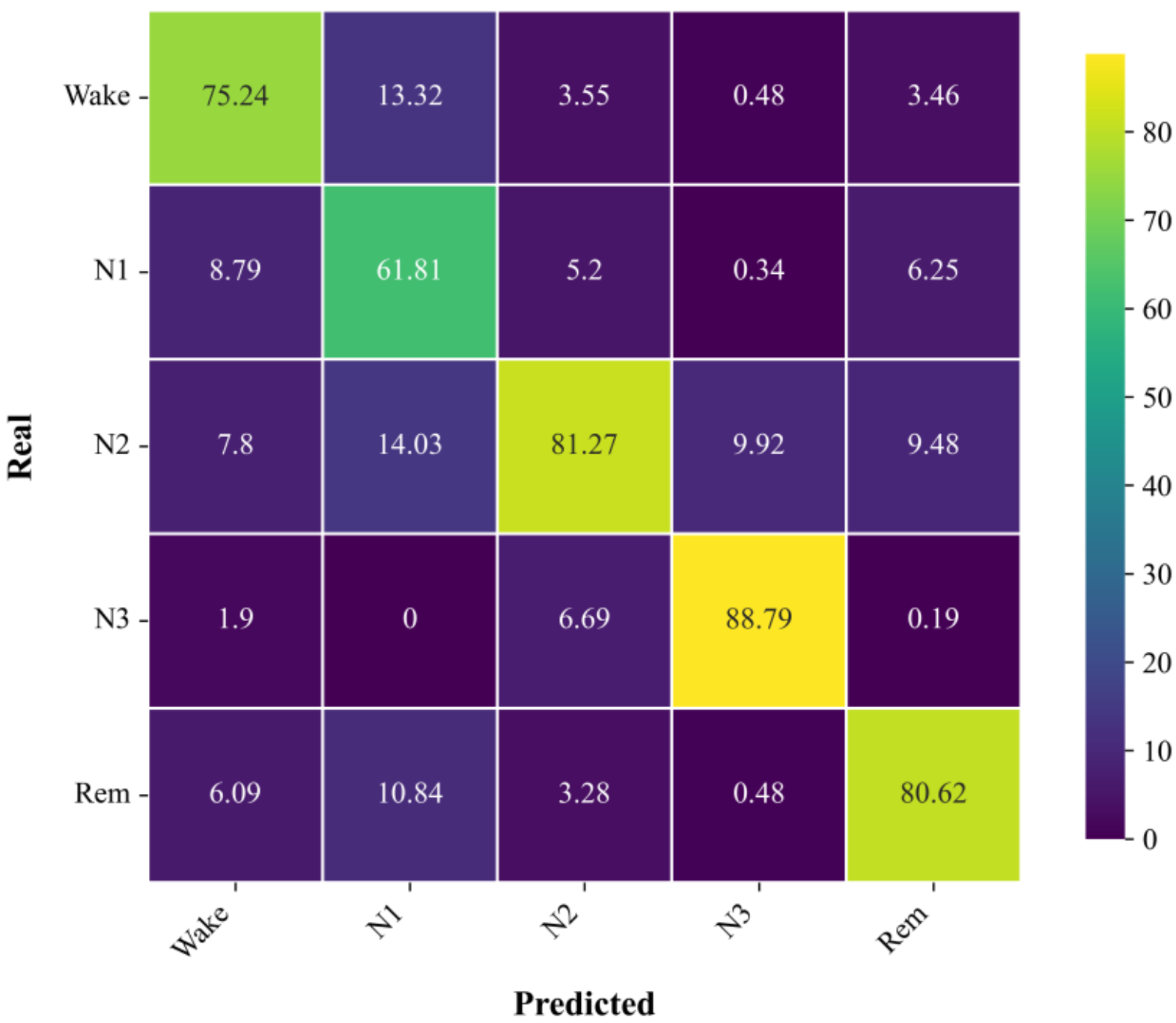

**Figure 7.** Confusion-matrix values for scorer 4 in the DOD-H dataset. Rows represent real stages and columns represent predicted stages.

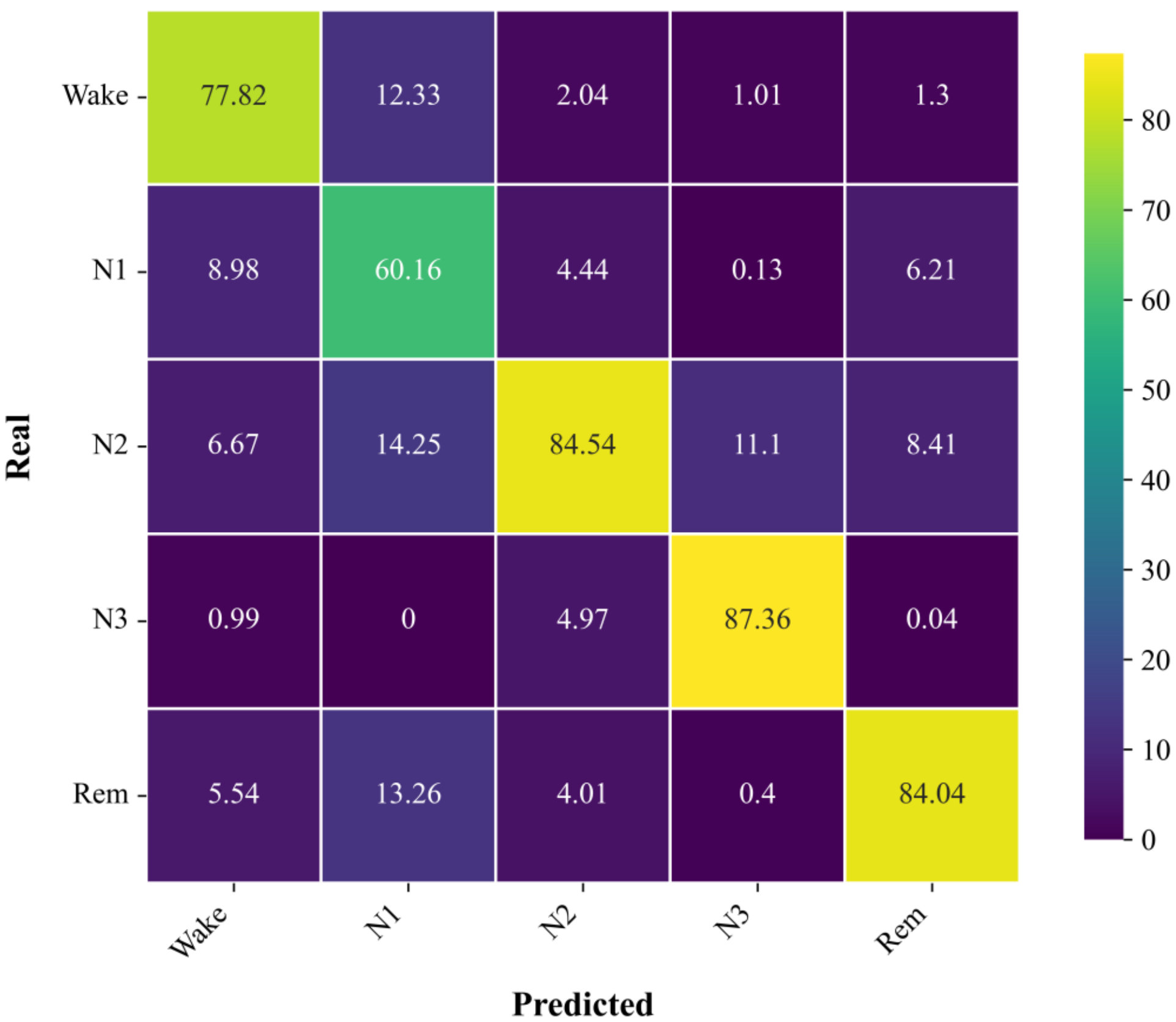

**Figure 8.** Confusion-matrix values for scorer 5 in the DOD-H dataset. Rows represent real stages and columns represent predicted stages.

### 3.6. Classifiers and Evaluation Metrics

The proposed hypnogram was evaluated using three standard classifiers: random forest (RF), support vector machine (SVM), and multilayer perceptron (MLP). For random forest, the number of trees was set to 300 and the remaining parameters were kept at their default values. For SVM, a Gaussian/RBF kernel was used and other parameters were kept at default values. The MLP consisted of three hidden layers with 50 neurons in each layer, a fixed random state for reproducibility, and a maximum of 200 training iterations; other parameters were kept at default values.

Each classifier was evaluated using EEG-only features and combined EEG+EMG features. The reported metrics include accuracy, precision, total F1-score, and class-specific F1-scores for W, N1, N2, N3, and REM. The same evaluation design was applied to BSH, DH, and LBH so that differences in performance reflect the label source rather than changes in the classifier setup.

$$Accuracy = \frac{TP + TN}{TP + TN + FP + FN} \quad (5)$$

$$precesion = \frac{TP}{TP + FP} \quad (6)$$

$$F1 = \frac{2 \times precision \times recall}{precision + recall} \quad (7)$$

## 4. Experiments and Results

This section reports the classification results for the three hypnograms. BSH denotes the best-scorer hypnogram, DH denotes the hypnogram distributed with the dataset, and LBH denotes the proposed learning-based hypnogram. All values are reported as percentages.

### 4.1. Random Forest Results

Random forest achieved the strongest overall performance among the three classifiers. In both datasets and both signal configurations, LBH produced higher total accuracy than BSH and DH. The highest scores were obtained when EEG and EMG features were combined.

**Table 5.** Random forest overall classification results on DOD-H and DOD-O.

| Dataset | Signal(s) | Hypnogram | Accuracy (%) | Precision (%) | Overall F1 (%) |
|---|---|---|---|---|---|
| DOD-H | EEG | BSH | 81.53 | 79.90 | 79.74 |
| DOD-H | EEG | DH | 82.85 | 81.56 | 81.57 |
| DOD-H | EEG | LBH | 83.87 | 82.89 | 82.80 |
| DOD-H | EEG+EMG | BSH | 83.56 | 82.46 | 82.19 |
| DOD-H | EEG+EMG | DH | 84.95 | 84.18 | 83.97 |
| DOD-H | EEG+EMG | LBH | 86.07 | 85.46 | 85.29 |
| DOD-O | EEG | BSH | 79.67 | 78.14 | 78.14 |
| DOD-O | EEG | DH | 80.73 | 79.32 | 78.80 |
| DOD-O | EEG | LBH | 82.29 | 81.22 | 80.90 |
| DOD-O | EEG+EMG | BSH | 83.66 | 81.73 | 82.14 |
| DOD-O | EEG+EMG | DH | 84.48 | 83.49 | 82.76 |
| DOD-O | EEG+EMG | LBH | 86.04 | 85.21 | 84.70 |

**Table 6.** Random forest class-specific F1-scores on DOD-H and DOD-O.

| Dataset | Signal(s) | Hypnogram | W | N1 | N2 | N3 | REM |
|---|---|---|---|---|---|---|---|
| DOD-H | EEG | BSH | 81.17 | 20.71 | 87.55 | 81.12 | 80.44 |
| DOD-H | EEG | DH | 83.16 | 23.95 | 88.10 | 84.77 | 80.08 |
| DOD-H | EEG | LBH | 82.91 | 20.51 | 88.65 | 85.42 | 79.85 |
| DOD-H | EEG+EMG | BSH | 82.02 | 30.32 | 88.57 | 84.16 | 84.57 |
| DOD-H | EEG+EMG | DH | 83.33 | 34.27 | 89.33 | 86.13 | 85.05 |
| DOD-H | EEG+EMG | LBH | 83.55 | 31.89 | 90.04 | 87.53 | 84.63 |
| DOD-O | EEG | BSH | 84.99 | 0.50 | 84.08 | 68.01 | 70.40 |
| DOD-O | EEG | DH | 84.34 | 6.78 | 85.99 | 80.54 | 72.56 |
| DOD-O | EEG | LBH | 84.40 | 3.80 | 87.40 | 77.08 | 72.85 |
| DOD-O | EEG+EMG | BSH | 87.04 | 0.12 | 86.64 | 73.53 | 82.03 |
| DOD-O | EEG+EMG | DH | 86.36 | 14.10 | 88.19 | 84.73 | 83.40 |
| DOD-O | EEG+EMG | LBH | 86.57 | 6.54 | 89.55 | 82.47 | 83.43 |

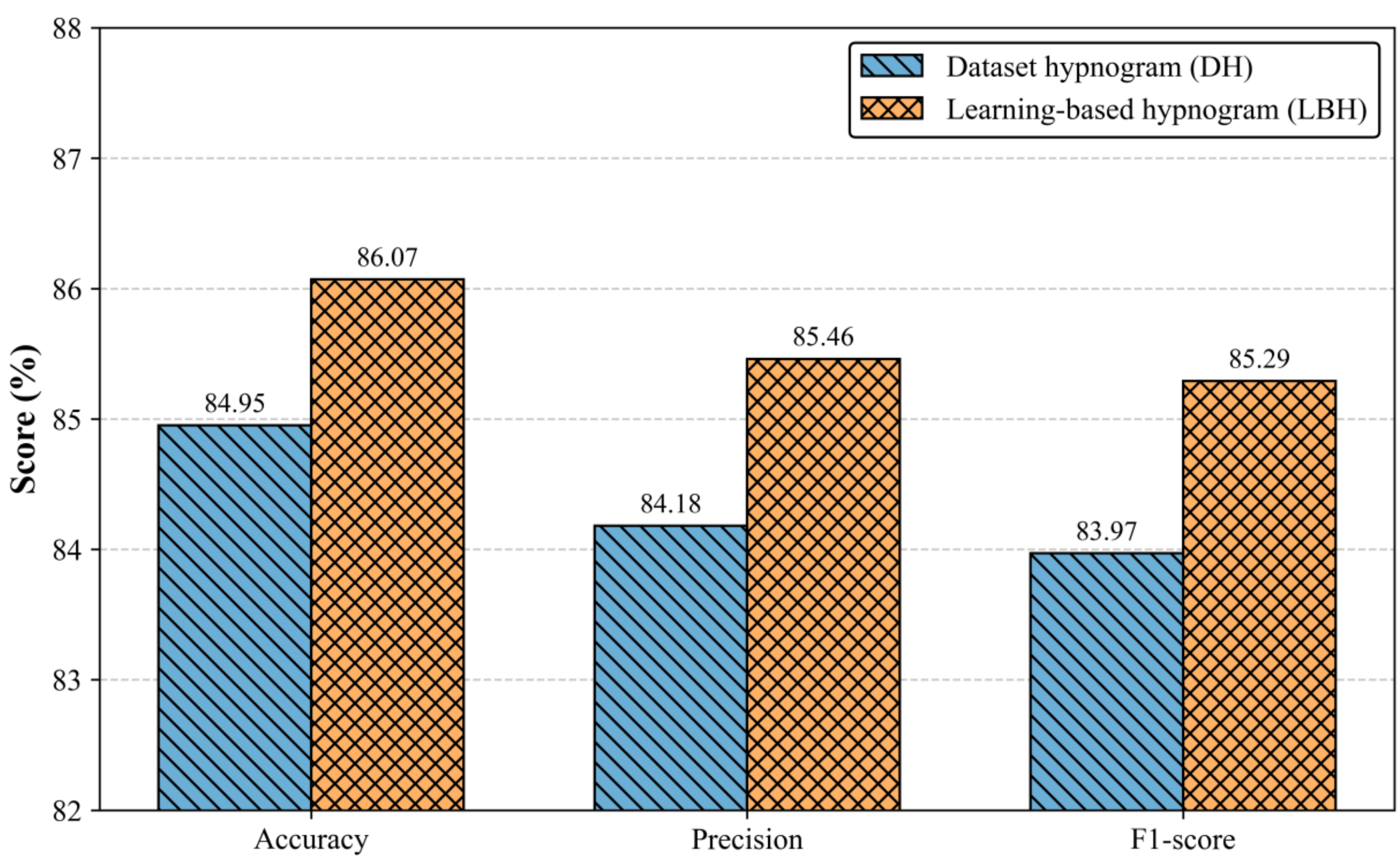

**Figure 9.** Random forest results using EEG+EMG features with DH and LBH on the DOD-H dataset.

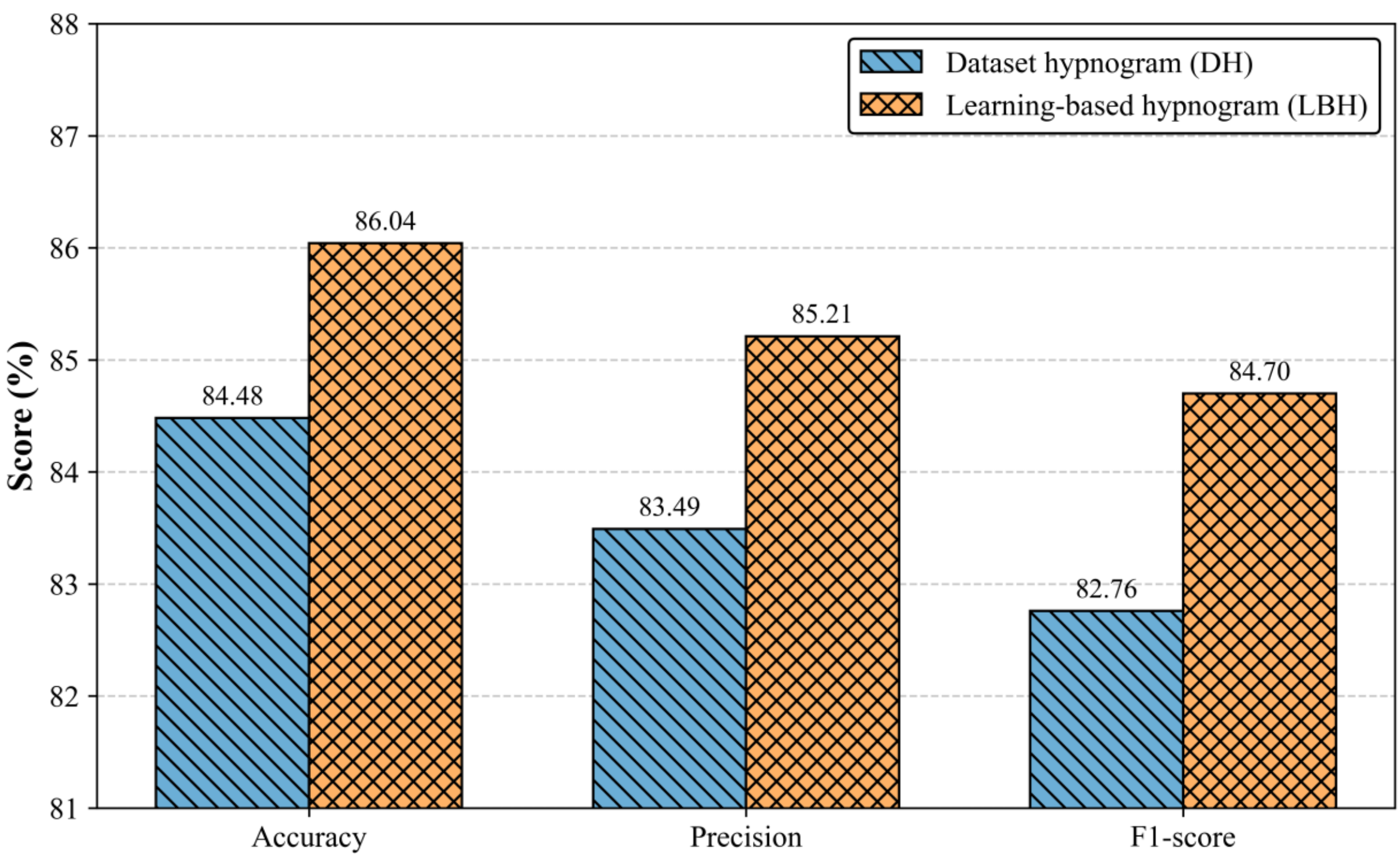

**Figure 10.** Random forest results using EEG+EMG features with DH and LBH on the DOD-O dataset.

## 4.2. Support Vector Machine Results

The SVM experiments show the same general pattern as the random forest experiments. LBH improved the overall accuracy, precision, and F1-score relative to DH and BSH in the EEG-only and EEG+EMG settings for both datasets. In DOD-O with EEG+EMG features, SVM accuracy increased from 81.32% with DH to 83.01% with LBH.

**Table 7.** Support vector machine overall classification results on DOD-H and DOD-O.

| Dataset | Signal(s) | Hypnogram | Accuracy (%) | Precision (%) | Overall F1 (%) |
|---|---|---|---|---|---|
| DOD-H | EEG | BSH | 79.11 | 77.11 | 77.28 |
| DOD-H | EEG | DH | 80.21 | 78.70 | 78.85 |
| DOD-H | EEG | LBH | 81.47 | 80.47 | 80.27 |
| DOD-H | EEG+EMG | BSH | 79.64 | 77.65 | 77.89 |
| DOD-H | EEG+EMG | DH | 81.18 | 79.79 | 79.94 |
| DOD-H | EEG+EMG | LBH | 82.66 | 81.65 | 81.69 |
| DOD-O | EEG | BSH | 76.40 | 76.64 | 75.04 |
| DOD-O | EEG | DH | 77.80 | 75.20 | 75.74 |
| DOD-O | EEG | LBH | 79.20 | 76.48 | 77.73 |
| DOD-O | EEG+EMG | BSH | 79.82 | 78.48 | 78.69 |
| DOD-O | EEG+EMG | DH | 81.32 | 78.92 | 79.25 |
| DOD-O | EEG+EMG | LBH | 83.01 | 80.96 | 81.55 |

**Table 8.** Support vector machine class-specific F1-scores on DOD-H and DOD-O.

| Dataset | Signal(s) | Hypnogram | W | N1 | N2 | N3 | REM |
|---|---|---|---|---|---|---|---|
| DOD-H | EEG | BSH | 76.05 | 18.16 | 85.64 | 76.59 | 79.71 |
| DOD-H | EEG | DH | 78.62 | 22.50 | 85.80 | 78.95 | 79.38 |
| DOD-H | EEG | LBH | 78.35 | 17.91 | 86.63 | 80.05 | 79.36 |
| DOD-H | EEG+EMG | BSH | 78.01 | 18.01 | 86.45 | 78.02 | 78.59 |
| DOD-H | EEG+EMG | DH | 80.21 | 21.87 | 87.13 | 81.78 | 78.72 |
| DOD-H | EEG+EMG | LBH | 80.51 | 20.53 | 87.90 | 84.01 | 78.52 |
| DOD-O | EEG | BSH | 88.33 | 2.81 | 84.94 | 77.92 | 67.25 |
| DOD-O | EEG | DH | 81.34 | 1.45 | 84.13 | 77.12 | 66.54 |
| DOD-O | EEG | LBH | 81.63 | 0 | 85.37 | 70.85 | 66.96 |
| DOD-O | EEG+EMG | BSH | 82.05 | 3.56 | 86.19 | 77.77 | 80.47 |
| DOD-O | EEG+EMG | DH | 83.42 | 3.55 | 85.83 | 78.93 | 79.48 |
| DOD-O | EEG+EMG | LBH | 83.79 | 0.74 | 87.32 | 78.74 | 80.09 |

## 4.3. Multilayer Perceptron Results

The MLP classifier also benefited from the proposed hypnogram. Although its absolute scores were lower than the random forest scores, LBH consistently improved total performance across the reported experiments. In DOD-O with EEG+EMG features, MLP accuracy increased from 80.95% with DH to 82.52% with LBH, and the total F1-score increased from 79.19% to 81.34%.

**Table 9.** Multilayer perceptron overall classification results on DOD-H and DOD-O.

| Dataset | Signal(s) | Hypnogram | Accuracy (%) | Precision (%) | Overall F1 (%) |
|---|---|---|---|---|---|
| DOD-H | EEG | BSH | 77.38 | 74.21 | 75.08 |
| DOD-H | EEG | DH | 78.83 | 77.35 | 77.05 |
| DOD-H | EEG | LBH | 79.70 | 77.93 | 78.15 |
| DOD-H | EEG+EMG | BSH | 78.31 | 77.73 | 77.09 |
| DOD-H | EEG+EMG | DH | 79.12 | 78.67 | 78.36 |
| DOD-H | EEG+EMG | LBH | 80.32 | 80.26 | 79.60 |
| DOD-O | EEG | BSH | 76.83 | 74.43 | 75.43 |
| DOD-O | EEG | DH | 78.29 | 76.60 | 76.54 |
| DOD-O | EEG | LBH | 79.90 | 78.33 | 78.48 |
| DOD-O | EEG+EMG | BSH | 79.50 | 77.57 | 78.08 |
| DOD-O | EEG+EMG | DH | 80.95 | 79.53 | 79.19 |
| DOD-O | EEG+EMG | LBH | 82.52 | 81.45 | 81.34 |

**Table 10.** Multilayer perceptron class-specific F1-scores on DOD-H and DOD-O.

| Dataset | Signal(s) | Hypnogram | W | N1 | N2 | N3 | REM |
|---|---|---|---|---|---|---|---|
| DOD-H | EEG | BSH | 76.26 | 8 | 84.58 | 75.11 | 75.55 |
| DOD-H | EEG | DH | 78.96 | 12.15 | 84.90 | 77.81 | 76.19 |
| DOD-H | EEG | LBH | 76.12 | 4.82 | 85.78 | 81.09 | 74.20 |
| DOD-H | EEG+EMG | BSH | 71.90 | 24.82 | 85.07 | 78.39 | 79.53 |
| DOD-H | EEG+EMG | DH | 71.33 | 27.72 | 85.27 | 81.85 | 79.01 |
| DOD-H | EEG+EMG | LBH | 71.56 | 23.60 | 86.08 | 83.28 | 78.02 |
| DOD-O | EEG | BSH | 83.10 | 0.11 | 81.59 | 65.81 | 65 |
| DOD-O | EEG | DH | 82.95 | 5.85 | 84.11 | 76.16 | 68.95 |
| DOD-O | EEG | LBH | 82.58 | 2.12 | 85.80 | 71.62 | 68.29 |
| DOD-O | EEG+EMG | BSH | 83.24 | 0.22 | 83.28 | 66.18 | 76.41 |
| DOD-O | EEG+EMG | DH | 83.29 | 9.51 | 85.45 | 78.65 | 78.58 |
| DOD-O | EEG+EMG | LBH | 83.90 | 6.67 | 86.78 | 76.79 | 78.07 |

### 4.4. Main Observations

- LBH improved the total accuracy, precision, and F1-score in every classifier, dataset, and signal configuration reported in the experiments.
- The best overall performance was obtained by random forest with EEG+EMG features and LBH: 86.07% accuracy on DOD-H and 86.04% accuracy on DOD-O.
- Adding chin EMG features generally improved performance relative to EEG-only features, especially for the random forest classifier.
- N1 remained the most difficult stage across datasets and classifiers. This is consistent with the agreement analysis, where N1 showed the lowest full-scorer agreement and the highest proportion of three-scorer agreement.
- The proposed method improves the label-generation process itself and is therefore not tied to a single final classifier architecture.

## 5. Discussion

The results indicate that the proposed personalized scorer modeling strategy provides a more effective reference hypnogram than the two baselines evaluated in this study. The improvement is observed not only for the classifier used to generate the scorer profiles but also for SVM and MLP classifiers, suggesting that LBH captures useful label information rather than simply overfitting to a single model.

The central advantage of LBH is that it treats scorer reliability as stage-dependent. In manual sleep staging, a scorer may be highly consistent for Wake or N2 while being less reliable for N1 or stage transitions. A single global scorer score cannot represent these differences. By normalizing confusion-matrix columns, the proposed method estimates how much support each scorer's assigned label gives to every candidate sleep stage. This allows all scorers to contribute while reducing the influence of scorer-stage combinations that are less consistent with the physiological feature patterns.

The results also show the importance of multimodal information. EEG features contain substantial information about sleep-stage transitions, spindles, slow waves, and REM-related activity. Chin EMG adds complementary information, particularly for distinguishing REM, Wake, and muscle-tone-related patterns. The EEG+EMG setting produced the strongest random forest results in both datasets.

The low F1-scores for N1 should not be hidden because they provide important information about the difficulty of this task. N1 is a transitional and low-prevalence stage, and the agreement tables show that scorers disagree most often in this class. The proposed LBH improves total metrics, but N1 remains challenging. This finding supports the need for label-generation methods that explicitly model uncertainty and scorer-specific behavior.

Compared with approaches that focus only on improving classifier complexity, this study addresses the quality of the target label. In multi-scored datasets, the reference hypnogram is not a neutral object; it is produced by an aggregation rule. Improving this rule can improve model training and evaluation even before changing the classifier architecture.

## 6. Limitations and Future Work

- The proposed scorer profiles were derived from machine-learning confusion matrices. These profiles quantify model-derived consistency between features and scorer labels, but they are not direct clinical proof that one scorer is correct and another is incorrect.
- Only the C3-M2 EEG channel and chin EMG were used. Future work can evaluate whether adding EOG, respiratory, ECG, or additional EEG channels further improves the generated hypnogram.

- The current experiments use conventional time- and frequency-domain features. Deep representation learning could be used in future work to generate scorer profiles from raw signals or time-frequency representations.
- N1 classification remains weak. Future work should explicitly address class imbalance, transitional epochs, and uncertainty-aware evaluation for N1.
- A subject-independent validation protocol and external validation on additional multi-scored datasets would further strengthen the generalizability of the proposed label-generation method.

## 7. Conclusion

This study proposed a personalized scorer modeling framework for generating learning-based hypnograms from multi-scored sleep datasets. Instead of relying only on a single scorer or a majority-based dataset hypnogram, the proposed method models the stage-specific behavior of each scorer using confusion matrices and aggregates the resulting probabilities to select the final sleep-stage label. Experiments on DOD-H and DOD-O showed that the proposed LBH improves accuracy, precision, and F1-score across random forest, SVM, and MLP classifiers and under EEG-only and EEG+EMG feature configurations. The best results were obtained with random forest and EEG+EMG features, reaching 86.07% accuracy on DOD-H and 86.04% accuracy on DOD-O. The findings suggest that improving the construction of the target hypnogram is a valuable and complementary direction for automatic sleep staging research.

## Acknowledgement

The authors declare that no funds, grants, or other support were received during the preparation of this manuscript.

## Data and Code Availability

The hypnograms proposed in this study and the hypnograms provided by the five scorers for both DOD-H and DOD-O are publicly available at *https://github.com/Edris-Hosseini/dreem-learning-hypnogram.*

## Conflict of Interest

The authors declare no competing interests.

## Ethics Statement

This study used publicly available, de-identified datasets. No new human-subject data were collected for this work.

## AI Acknowledgement

The authors confirm that no artificial intelligence (AI) tools were used in the conceptualization, design, data collection, analysis, or interpretation of this study. Limited AI-based software assistance was employed solely for grammar and language editing purposes but was fully humanized.